# A Machine Learning Approach for Modelling Parking Duration in Urban Land-use

Janak Parmar [a], Pritikana Das [b], Sanjaykumar Dave [c]

[a] Transportation Engineering and Planning Section, Department of Civil Engineering, Sardar Vallabhbhai National Institute of Technology (SVNIT), Surat, 395007, India.
[b] Civil Engineering Department, Maulana Azad National Institute of Technology, Bhopal, 462003, India.
[c] Civil Engineering Department, The Maharaja Sayajirao University of Baroda, Vadodara, 390001, India.

**Abstract:** Parking is an inevitable issue in the fast-growing developing countries. Increasing number of vehicles require more and more urban land to be allocated for parking. However, a little attention has been conferred to the parking issues in developing countries like India. This study proposes a model for analysing the influence of car user's socioeconomic and travel characteristics on parking duration. Specifically, artificial neural network (ANN) is deployed to capture the interrelationship between driver's characteristics and parking duration. ANNs are highly efficient in learning and recognizing connections between parameters for best prediction of an outcome. Since, utility of ANNs has been critically limited due to its 'Black Box' nature, the study involves the use of Garson's algorithm and Local interpretable model-agnostic explanations (LIME) for model interpretations. LIME shows the prediction for any classification, by approximating it locally with the developed interpretable model. This study is based on microdata collected on-site through interview surveys considering two land-uses: office-business and market/shopping. Results revealed the higher probability of prediction through LIME and therefore, the methodology can be adopted ubiquitously. Further, the policy implications are discussed based on the results for both land-uses. This unique study could lead to enhanced parking policy and management to achieve the sustainability goals.

**Corresponding Author:** mailjanakp@gmail.com

## 1. Introduction

In the modern era, population and economic growth in conjunction with an increasing living standard of the people are to blame for rising number of automobiles. Ever-increasing number of private vehicles coupled with an inefficient public transit system increasingly extracting the urban land for parking allocation which otherwise would overload the curb parking. Subsequently, it would reduce the traffic safety, air quality (reduced carriageway tends to increase congestion), etc. In critical phenomenon, when vehicle volume exceeds the network capacity, traffic congestion arises which has negative impacts on social as well as economic life of the society as a whole. As per the study by Boston Consulting Group (April 2018), leading consequences like delay and vehicle idling culminate into huge economic loss, estimated as USD 22 billion per year in four major cities Delhi, Kolkata, Mumbai and Bangalore in India. According to the Ministry of Road Transport and Highways, India (MoRT&H), there is whooping growth in motor vehicle population of nearly 460% from 55 million in 2001 to 253.3 million in 2017 in India. The need for parking space is directly in nexus of demand associated with the upsurge in the vehicle plying on the roads. The challenges for parking are exacerbated as the parking supply in general is determined based on the parking necessity, lack in consideration of direct and indirect cost, now and in future as the parking spaces are built and convert the valuable lands to absorb the requirements (Shoup and Pickrell, 1978; Tumlin, 2012). In India, Delhi has the maximum percentage of its land under roads and about 14% of the road length is used for parking (Roychowdhury et. al, 2018). In Jaipur, this share is 56% while in many other big cities it is more than 40%, which is not only a wasteful use of land, but also undermines the utility of public transit. It has been estimated that the annual demand for car parking space in Delhi can be equivalent to as much as 471 football fields. A highly accurate prediction of duration model will impart great knowledge to planners and policy makers in framing efficient strategies to manage available parking supply and to improve the patronage of public transport infrastructure.

Parking demand refers to the amount of parking that would be used at a certain time, place and duration. It is a function of vehicle ownership, trip rate, mode choice, duration, location and land-use, etc. Parking demand has usually daily, weekly and even annual cycles. For example, parking demand peaks on weekdays at office-business areas while on weekends at market, restaurants, theatres, etc. Moreover, parking demand can change with transportation, land-use and demographic patterns. If a building change from residential to industrial or office-use, density and neighbourhood pattern may change, quality of transit service may change, all of them affect the parking demand and duration. Different types of trip purpose have different parking duration. Commuters need parking for larger duration and hence are relatively more price sensitive. In regards to above, there is a requirement of efficient parking management strategies at policy level especially in Central Business Districts (CBDs) where parking is limited and demand is intense. A great amount of parking demand in these CBDs is generated by the visitors who park their car for specified time period (called parking duration) before departing to their origin place. An effective management of the parking demand especially in large central cities is an ongoing challenge as they face competing objectives and ever-increasing demands for space (De Cerreño, 2004). At this juncture, parking has become one of the major concerns for transport planners in the scenario of perennially burgeoning private vehicle ownership. Hence, at planning and development stage, it is inevitable to address the infrastructural and sometimes technological demands which has limited resources to supply. As the main objective of the parking management is to balance the demand and supply, this study emphasized to model a parking duration in relation with the driver's socioeconomic and travel characteristics for two land-uses i.e., office-business (OB) and market/shopping (MS), to discover the parameters to be targeted at a strategic level of planning.

Various conventional regression models (e.g., MLR, MNL, GLM, etc.) are widely used for the statistical analysis requiring crisp input parameters. However, these models are not capable to recognize any data patterns and complex relations between the variables (Warner and Misra, 1996). Artificial Intelligence (AI) methods, especially those based on machine learning, are becoming essential for analysis where there is complexity in data and for decision support in various research areas including traffic and transportation. Artificial neural networks (ANNs) are inherently non-linear and non-parametric in nature with sets of neurons that can be used for resembling the association of input and output signals of complex system. Though ANNs are termed as powerful tools providing greater accuracy compared to conventional models, they often criticized as 'black box' as the results are difficult to interpret and the synaptic weight matrix containing the knowledge of trained network cannot be easily decrypted. The black box issue in the sense that the model provided by the trained network will not deliver insight into the form of the function as in case of conventional models because of non-linear relationship between the network weights and the target variable/s. In contrast to the generalized regression models which generate reproducible regression coefficients and clinical form of modeled function, multiple neural networks with different weights generate similar kind of predictions for a same training dataset and network topology which create confusions.

Recognizing the discussed issue, many methods have been developed to understand the function of ANN (Bach et al., 2015; Friedman, 2001; Garson, 1991; Lek et al., 1996; Ribeiro et al., 2016b). This article uses two methods namely Garson's algorithm (Garson, 1991) and local interpretable model-agnostic explanations (LIME) (Ribeiro et al., 2016b) for the interpretation of trained neural network. Former method estimates the magnitude of relative importance of the predictors to the target variable by dissecting model weights while a later one developed recently, shows the prediction of any classification or regression, by approximating it locally with the developed interpretable model. Different library packages available with R have been deployed for the analysis and interpretation. The method and developed model help in planning and policy-implications based on the derived results.

## 2. Literature Review

Multiple studies have been done in past focusing the parking concerns. In general, parking models are distinguished in parking design models, parking allocation models, parking search models, parking choice models and parking interaction models (Young, 2008). This study aims to model parking duration which is basically a type of interaction model. Parking interaction models are used for the investigation of response- interaction between users and parking facilities, new policies and applications. Parking demand in general influenced by whole prospect of social, economic and environmental factors, some of them are travel distance and travel time (Hensher and King, 2001; Sen et al., 2016), searching time for parking space and walk time to destination (Hilvert et al., 2012; Lau et al., 2005; Tong et al., 2013), parking price (Hensher and King, 2001; Lim et al., 2017; Tiexin et al., 2012), land-use characteristics (Bai et al., 2011; Fiez and Ratliff, 2019; Lau et al., 2005; Wong et al., 2000) and socioeconomic characteristics of driver (Aderamo and Salau, 2013; Ghuzlan et al., 2016). Parking demand in any area is generally a function of land-use characteristics of that area. About sixty-nine different land use classifications are available as per The Parking Generation Manual of the Institute of Transportation Engineers (ITE) . Each statistics of land use characters more or less influence the parking generation for the particular building/area. Parking generation rate is the demand for the parking space generated from a given land use in *per unit area* (Tiexin et al., 2012). Kefei established a static parking generation rate model considering land-use characteristics which was as per equation 1 (Kefei, 1994). Tiexin (Tiexin et al., 2012) improved a parking generation rate model by considering several factors such as average turnover rate, parking lot occupancy, parking price impact coefficient, level of service (LOS), vehicle growth factor, etc. (equation 2). In that, LOS was taken arbitrarily without considering factors those affect the LOS.

$$y = \sum_{i=1}^{n} a_i \times R_i;\ (i = 1,2,\dots,n) \qquad (1)$$

$$y = \sum_{i=1}^{n} \frac{a_i \times R_i}{\mu_i \times \gamma_i} \times \delta \times L \times \beta;\ (i = 1,2,\dots,n) \qquad (2)$$

Where: $y$ refers to parking demand in terms of lot/space; $a_i$ refers to parking generation rate that is demand per unit area; $R_i$ is area of building in $m^2$; $\mu_i$ refers to average parking turnover rate; $\gamma_i$ refers to parking lot occupancy; $\delta$ is a parking level of service; $L$ refers to parking price impact coefficient; $\beta$ refers to motor vehicle growth coefficient. Further, this model was expanded by Das et al. (Das et al., 2016) in that, they included cost factor: the travel cost by car with respect to that by the public transit, and utility-based coefficient showing preference of choosing car over transit. In addition, they empirically derived LOS using analytical hierarchy process (AHP).

Huayan et al. (SHANG et al., 2007) analysed parking demand at university campus for different of day by conducting in-out survey at university main entrance and exit. The study was mainly aimed at estimating parking demand, duration and turn-over for different times of a day. Wong et al. (Wong et al., 2000) developed parking demand models for private cars and goods vehicles by establishing the accumulation profiles. The study assumed that the parking activity associated with the individual land-use variables produce a unique parking accumulation profile. Subsequently, Tong et al. (Tong et al., 2004) proposed a usage related parking demand-supply equilibrium model. This model adopts a network equilibrium approach considering multiple vehicle classes and different types of parking facilities. The study gives insights of using accumulation profiles for the parking choice and parking demand models. Fiez and Ratliff (Fiez and Ratliff, 2019) considered the spatial and temporal properties of the parking demand from the paid parking transactions in Seattle to identify zones and time periods with similar demand using Gaussian mixture model (GMM). They carried out an analysis to improve the traditional parking pricing policies by suggesting where and when to administer pricing schemes in terms of zones encompassing group of block-faces and rate periods.

Various studies examine how different parameters affect the parking model. The parking price has been considered as one of most influencing factors for parking choice and mode choice which directly affect the transportation demand. Ottosson et al. (Ottosson et al., 2013) investigated how parking pricing affect several parking characteristics, such as turn-over rate, parking duration, and parking accumulation using automatic transaction data from the pay-and-park stations in Seattle. Hensher and King (Hensher and King, 2001) examined the role of parking pricing in the generation of parking demand. They used stated preference survey data– gleaned through private car users and public transit users– to catch individual's behaviour which varies with parking price and hours of operation at various locations within the CBD. Al-Sahili and Hamadneh (Al-Sahili and Hamadneh, 2016) studied the parking requirement for different land uses such as residential, hotel, office and commercial/shopping to confine trips and parking associated with the specific land use. They used morning and evening peak period survey data for eleven different land uses to establish the relationship of parking demand generation with these land uses. Ghuzlan et al. (Ghuzlan et al., 2016) investigated the impact of building characteristics and various socioeconomic characteristics of parking users on the parking demand particularly for residential land-use. They considered various building location class for instance, suburban, urban non-CBD, urban inner CBD and urban outer CBD.

Researchers have incorporated various AI techniques such as ANN, support vector machine and decision tree analysis in for solving transportation problems. Neural Networks have been widely used in the demand forecasting and traffic behaviour analysis, such as gap acceptance studies which efficaciously showing the advantages of using it. ANN forecasts the output variable by minimizing the error term representing the deviation between input and output using specified training algorithm and at random learning rate (Black, 1995; Zhang et al., 1998). Hornik (1991) and Cybenko (1989) proved that back-propagation neural network (BPNN) with one hidden layer can approximate any continuous function with desired degree of precision, provided it contains sufficient number of nodes in the hidden layer. Ravi Sekhar et al. (Chalumuri et al., 2009) studied the application of neural networks in mode-choice modeling for second-order metropolitan cities of India and they used Garson's algorithm of synaptic weights to measure the influence of input parameter on the mode-choice. Bai et al., 2011 have applied BPNN approach to forecast the parking demand for which, the influencing factors were obtained by principal component analysis. They carried out survey at 8 different types of land-uses to develop a neural network model and claimed that BPNN has the ability that makes the prediction results much easier to reflect the actual parking demand. Das et al. (Das et al., 2015) analysed the interrelationship between pedestrian flow parameters by deterministic and ANN approach. They derived macroscopic fundamental diagram based on ANN model and proved that ANN approach model imparts better future prediction accuracy compared to deterministic approach. Despite the superiority of neural networks over statistical model, the structural flexibility of neural network leads to neural networks and statistical models end up with same model (Karlaftis and Vlahogianni, 2011).

It can be inferred from the literature survey that substantial amount of research has been done on the parking modeling and effect of various pricing policies on it. However, to the best knowledge of authors, no study investigated the impact of user's personal and travel characteristics in addition to parking price on the characteristics of parking demand such as parking duration. In addition, the 'black box' nature of the ANN qualified its applications in practical problems. With this background, the objective of the study is to identify the interrelationship between user's characteristics and parking demand- termed as parking duration and further, to predict the parking demand using the ANN interpretation techniques such as synaptic weight algorithm and LIME.

The study is limited to business cum commercial and purely market/shopping land-use as they are predominant in attracting high volume of trips in medium to large metropolitan cities.

## 3. Mathematical Framework

In this study, multilayer feedforward BPNN with single hidden layer is considered for modeling parking duration. The parking duration is classified for grouping the dataset and hence, the problem is essentially a classification problem to the neural networks. For instance, the solution would lie in the $n^{th}$ classification out of total 'm' classifications according to the considered characteristics of an individual 'i'. The model is developed and analyzed with the help of *nnet* package (version 7.3-12) available in R (version 3.6.1) which can fit a single-hidden-layer neural network. In addition, the *caret* (Classification And REgression Training) package (version 6.0-84) containing a set of tools for building machine learning models in R, is used to train a BPNN model. The framing of BPNN was structured in three layers viz., input (IL), hidden (HL) and output (OL) layers having a set of interconnected artificial neurons (See figure 1). Once the group of neurons assigned with the datapoints through IL, the neurons in IL transmit the weighted data to HLs and randomly select the bias through HLs ($B_1$ in figure 1). When it estimates the net sum at the hidden node, an output response is provided at the node using a specified activation function with an error term represented as $B_2$ in figure 1. As there are two layers except input layer in the model, suppose that the number of nodes of hidden layer is $z_l$ , $z_s^{(l)}$ is the input from the node $s$ in hidden layer and $\bar{z}_s^{(l)}$ stands for the output node $s$ in hidden layer, then the formula 3 and 4 can be obtained which are used for formulating output. In the proposed model, logistic (sigmoid) function was used as a main activation function $f(\ldots)$, which as shown in the equation 4.

$$z_s^{(l)} = W_s^{(l)} \bar{z}^{(l-1)} = \sum_{k=1}^{n_{l-1}} w_{sk}^{(l)} \, \bar{z}_k^{(l-1)};$$

$$s = 1,2, \ldots , n_l; \; k = 1,2, \ldots , n_{l-1} \qquad (3)$$

$$\bar{z}_s^{(l)} = f\left(z_s^{(l)}\right); \quad s = 1,2, \ldots , n_l \qquad (4)$$

Where:
$W_s^{(l)}$ is connection weight
Sigmoid: $f(x) = 1/(1 + e^{-x})$

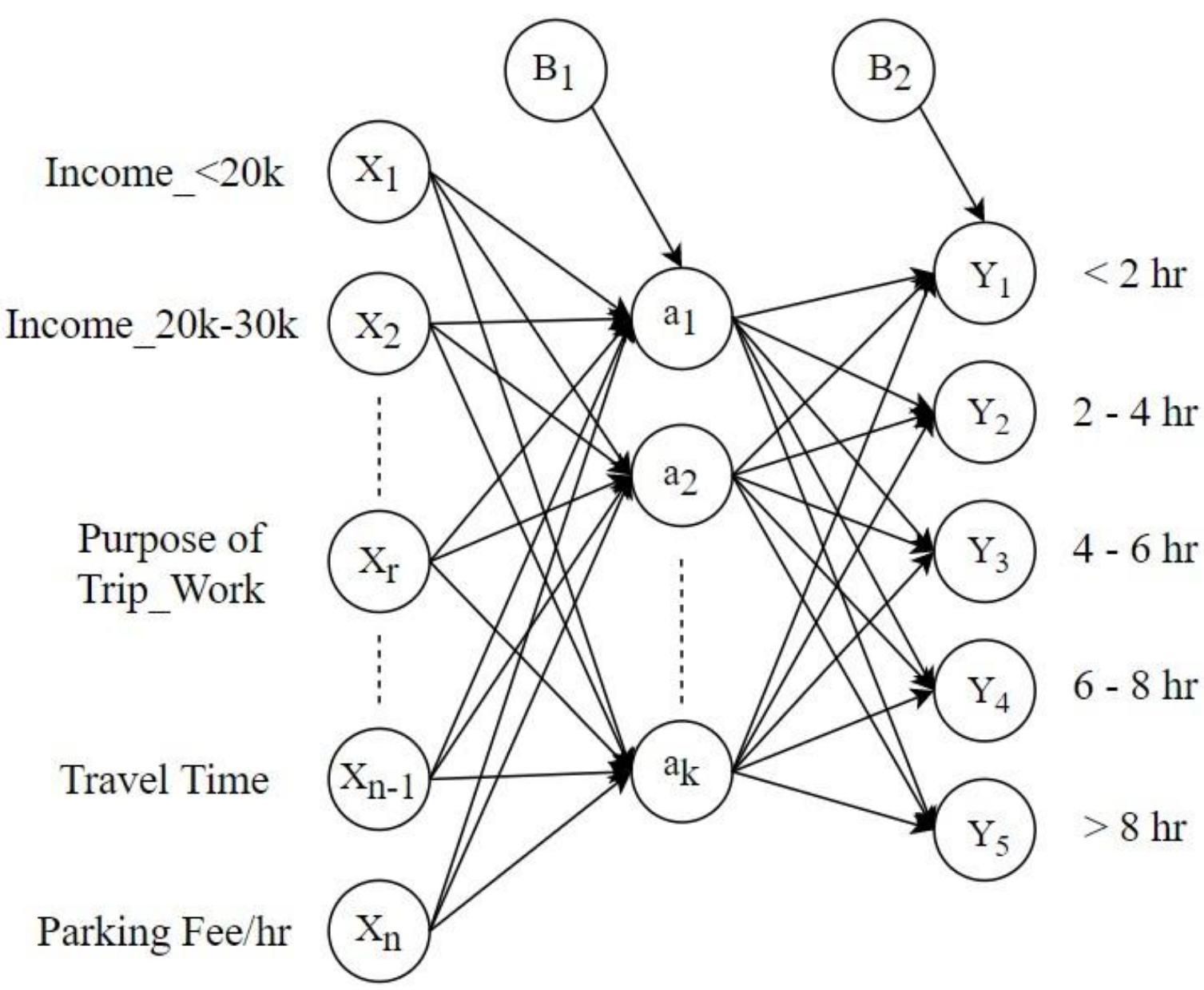

**Fig. 1.** Architecture of ANN Classifier

In model training, *k*-fold cross validation is used to select the best model with the tuning parameters. It creates *k* sets of datasets from the original dataset. A model is fit using the samples of *k*-1 datasets while a remaining 1 subset is used to validate the model and to estimate the performance measures. This process is repeated *k* times, with all observations are utilized for both training and testing, and each datapoint is used for testing exactly once. The two tuning parameters- *size* and *decay* are used to pick the setting associated with the best performance. The *size* parameter defines the number of hidden neurons in a single HL in the network, which are essentially the free parameters allow flexibility in model fitting. Higher the number of hidden neurons, greater the flexibility of the model but at the risk of over-fitting. Hence, the *decay* parameter regularizes the rate of decay for changing the weights those used by the back-propagation fitting algorithm. Following equation 5 denotes categorical cross-entropy (CE) loss function used in this model which is to be minimize in order to obtain an optimized neural net model.

$$CE = -\log\left(\frac{e^{z_s}}{\sum_s^C e^{z_s}}\right) + \lambda\left(\sum\nolimits_{s,k}\left(W_{sk}^{(l)}\right)^2 + \sum\nolimits_{s,k}\left(W_{sk}^{(l-1)}\right)^2\right) \tag{5}$$

where, $z_s$ is the output in the node *s* of the OL, *C* is the number of possible classification and *W* is the matrix of the link weights that is to be panelized by the *decay* λ.

For interpreting the obtained BPNN model results, two different algorithms viz. Garson's partitioning of weights algorithm and Local interpretable model-agnostic explanation (LIME) have been deployed. The weights connecting neurons in neural networks are partially analogous to parameter coefficients in standard regression model. The weights express the relative influence of information processed in the network such that input variables that have strong/weak negative/positive associations with the response variable. There are large number of adjustable weights in the network makes it flexible in modeling non-linear relationship but inflicts challenges for the interpretation. Garson (1991) proposed a method of partitioning connection weights to determine the relative importance of input variable within the network. He claimed that pooling and scaling all weights specific to a predictor generates a single value in the interval [0 1] that reflects relative predictor importance. The method has been later modified and harnessed by Goh (Goh, 1995). The details of the algorithm are given in equation 6. In R, relative importance can be estimated with the *NeuralNetTools* (1.5.2) package.

$$Q_{ik} = \frac{\sum_{j=1}^{L} |w_{ij}v_{jk}| \;/\; \sum_{r=1}^{N} |w_{rj}|}{\sum_{i=1}^{N}\sum_{j=1}^{L}\left(|w_{ij}v_{jk}| \;/\; \sum_{r=1}^{N} |w_{rj}|\right)} \qquad (6)$$

where $w_{ij}$ is the weight between $i^{\text{th}}$ input neuron and $j^{\text{th}}$ hidden neuron, and $v_{jk}$ is the weight between $j^{\text{th}}$ hidden neuron and $k^{\text{th}}$ output neuron. $\sum_{r=1}^{N} |w_{rj}|$ is the sum of the connection weights between $N$ input neurons and the hidden neuron $j$. $Q_{ik}$ represents the percent influence of the input parameter on the output. This algorithm has limited applications since it cannot depict whether the parameter is positively or negatively correlated with the output node like conventional classifiers. The LIME algorithm outperforms these limitations which is described in the subsequent paragraph.

Further, LIME- Local Interpretable Model-agnostic Explanations is an algorithm that can be used to explain the predictions of classifications or regressions models, by approximating it locally with an interpretable model (Ribeiro et al., 2016b). It explains the outcome of black box models by fitting a local model around the point under consideration and perturbations of this point. LIME identifies an interpretable black box model over the interpretable representation that is *locally faithful* to the classifier, by approximating it in the vicinity of a discrete instance. Let $x \in \mathbb{R}^d$ as the original representation of an instance that is to be explained, and $x' \in \mathbb{R}^{d'}$ be a vector for its interpretable representation. Formally, the explanation model is $g$: $\mathbb{R}^{d'} \rightarrow \mathbb{R}$, $g \in G$, where $G$ is class representing potentially interpretable models. As noted in (Ribeiro et al., 2016a), not every $g \in G$ is simple enough to be interpretable, letting introduction of $\Omega(g)$ as a measure of complexity (opposed to interpretability) of $g$, which may be either a soft constraint or a hard constraint.

Let $f$: $\mathbb{R}^d \rightarrow \mathbb{R}$ is the black box model being explained, for instance in classification $f(x)$ is the probability that $x$ belongs to certain class. Further, assume $\Pi_x(z)$ as a proximity measure between two individuals $x$ and $z$, so as to define the locality around $x$. Finally, introduce $\mathcal{L}(f, g, \Pi_x)$ as a measure to define the faithfulness of $g$ in approximating $f$ in the locality described as $\Pi_x$. To ensure both interpretability and local fidelity, the function $\mathcal{L}(f, g, \Pi_x)$ must be minimized while having $\Omega(g)$ be low enough to be interpretable by humans. Consequently, the explanation $\xi(x)$ yielded by LIME is obtained by solving:

$$\xi(x) = \underset{g \in G}{\operatorname{argmin}}\, \mathcal{L}(f, g, \Pi x) + \Omega(g) \qquad (7)$$

This equation 7 can be used with different explanation families $G$, fidelity functions, $\mathcal{L}$ and complexity measures $\Omega$. $\mathcal{L}$ can be estimated by generating perturbed samples around $x$. For the current study, the *lime* package (0.5.0) is used to implement LIME algorithm.

Currently, very limited literature available which studies the interpretability of the AI models particularly in the context of transportation engineering and planning. This proposed model can efficiently map the hidden features for finding the interrelationships between the considered parameters. Also, LIME is very efficient in retrieving this information *locally* from the black-box models. The subsequent sections describe the adaptability of this model using a case study.

## 4. Study Area and Data Collection

In the present study, NCT (National Capital Territory) of Delhi in India was chosen as the study area. Delhi has a population of 16.7 million with a population density of 11,297 / $km^2$ and about 97.5% of its total population living in urban area (2011). Public transit in Delhi has two major services viz. bus transport and metro rail. The intermediate para-transit (IPT) modes such as, autorickshaws (a 3-wheeler IPT), cycle-rickshaws and electric-rickshaws play a huge role in commuting as well as all other types of trips. DTC (Delhi Transport Corporation) operates approximately 5600 single-floor buses in mundane in addition to other cluster buses. Currently, Delhi Metro is the $7^{th}$ largest metro network in the world with the system length of 391.38 km and serving 285 stations. On an average, the combination of two serves almost 6.8 million daily ridership as per Economic Survey of Delhi 2018-19. Although having a sound public transit network, the use of personalized vehicles is very high in Delhi. It has 10.32 million of registered vehicles in terms of cars, jeeps and motor-cycles sharing 94% of the total vehicle population within NCT. Each of these vehicles require parking space at minimum of two places viz. at home and at work place for considerably high duration. This creates a huge demand for parking space in the city as stated earlier. Looking to this concern, this study attempts to analyse the parking demand in terms of duration to discover which parameters have most influence on it.

For the current study, data were collected through on-field interview survey using structured questionnaire at 13 different parking places within the city. These areas well represent the city characteristics as they attract the commute/shopping trips from all across the city because of the diversity at the selected destinations. The questionnaire encompassed the personal information, trip information and parking related information. This study mainly focused on parking at office-business oriented land-use (OBP) and market/shopping land-use (MSP). For OBP, 488 effective samples from 7 different parking lots were included for developing model. For MSP, 681 effective samples from 6 different parking lots were incorporated for model development. Moreover, it was estimated that around 25% of the OBP is employer-provided, that is the parking cost is subsidized or fully paid by the employers. The Fig. 2 portrays the locations and details of survey areas with number of parking lots surveyed (in brackets). The locations number 1 and 2 were selected for OBP while other four locations for MSP. At all these lots, parking charge is levied manually as no parking meter or other digital facilities available (like smart card in transit) to pay charges. These locations are well connected by means of city-bus and metro; however, they observe high demand of parking for private vehicles.

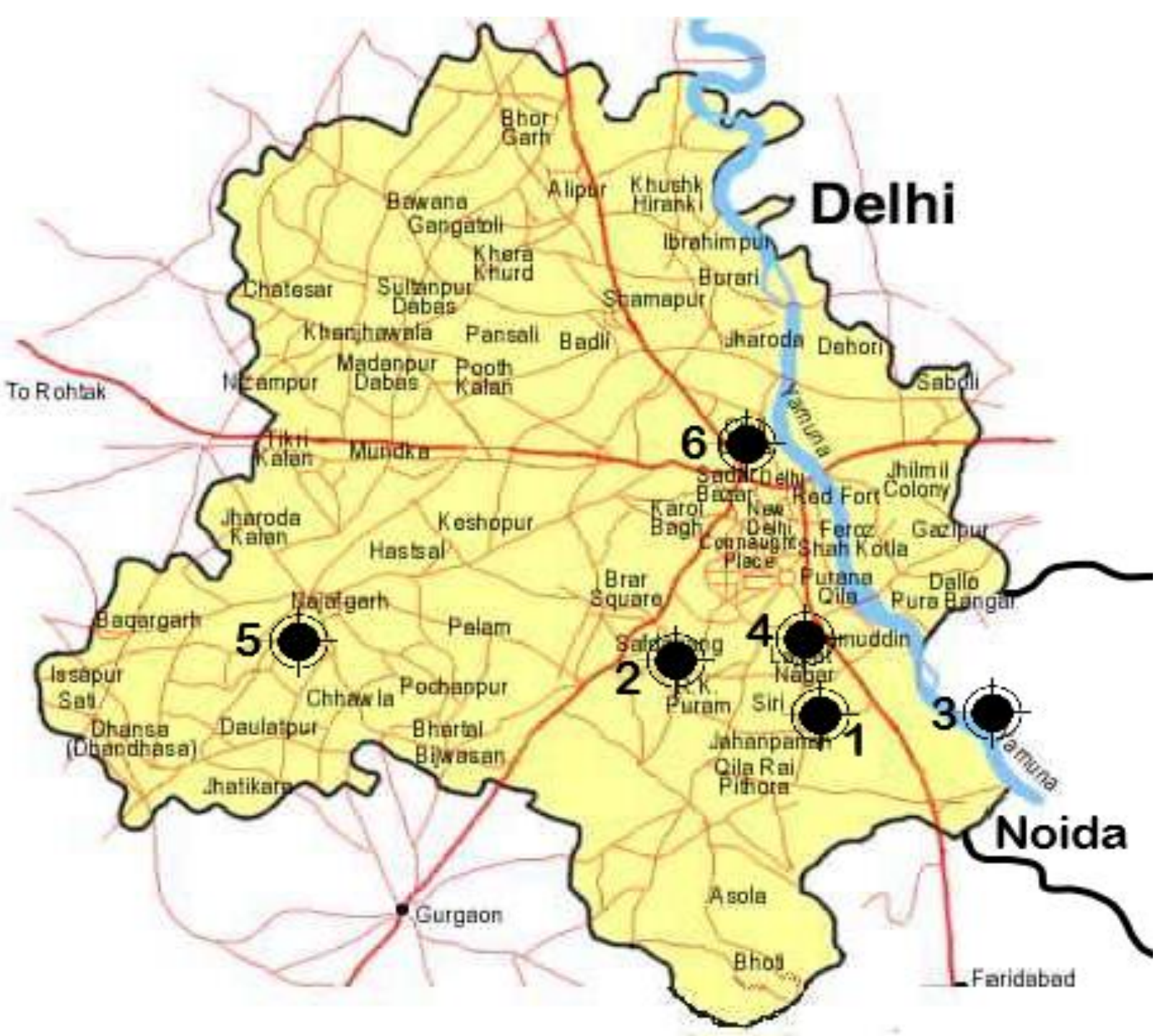

**Fig. 2.** Survey Locations in NCT of Delhi
**Legends: 1.** Nehru Place (3); **2.** Bhikaji Cama Place (4); **3.** Atta Market, Noida (2);
**4.** Lajpat Nagar (1); **5.** Dwarka Sec.12 (1); **6.** Kamla Nagar (2).

### 4.1. Data Briefing

The parameters considered in the model development include personal income, trip purpose, travel cost, travel time and average parking fees per hour (fph). In addition, the profession of driver was included in model particularly for OBP. Among selected variables, income, profession and trip purpose were considered as categorical variable while others were continuous type. The average parking fees (fph) were estimated based on the total paid parking charges and duration. Currently, the parking fees are 20 INR/hr with a maximum of 100 INR per day in Delhi. Table 1 shows the details of considered variables and their basic statistics. Parking duration (in hrs.) was classified into 5 groups: {<2, 2-4, 4-6, 6-8, >8}. The percentage distribution of the parking duration is represented in the Fig. 3. For OBP, analysis of parking duration distribution revealed that around 45% of cars were parked for more than 6 hrs while other 30% cars park for 2-6 hrs of duration.

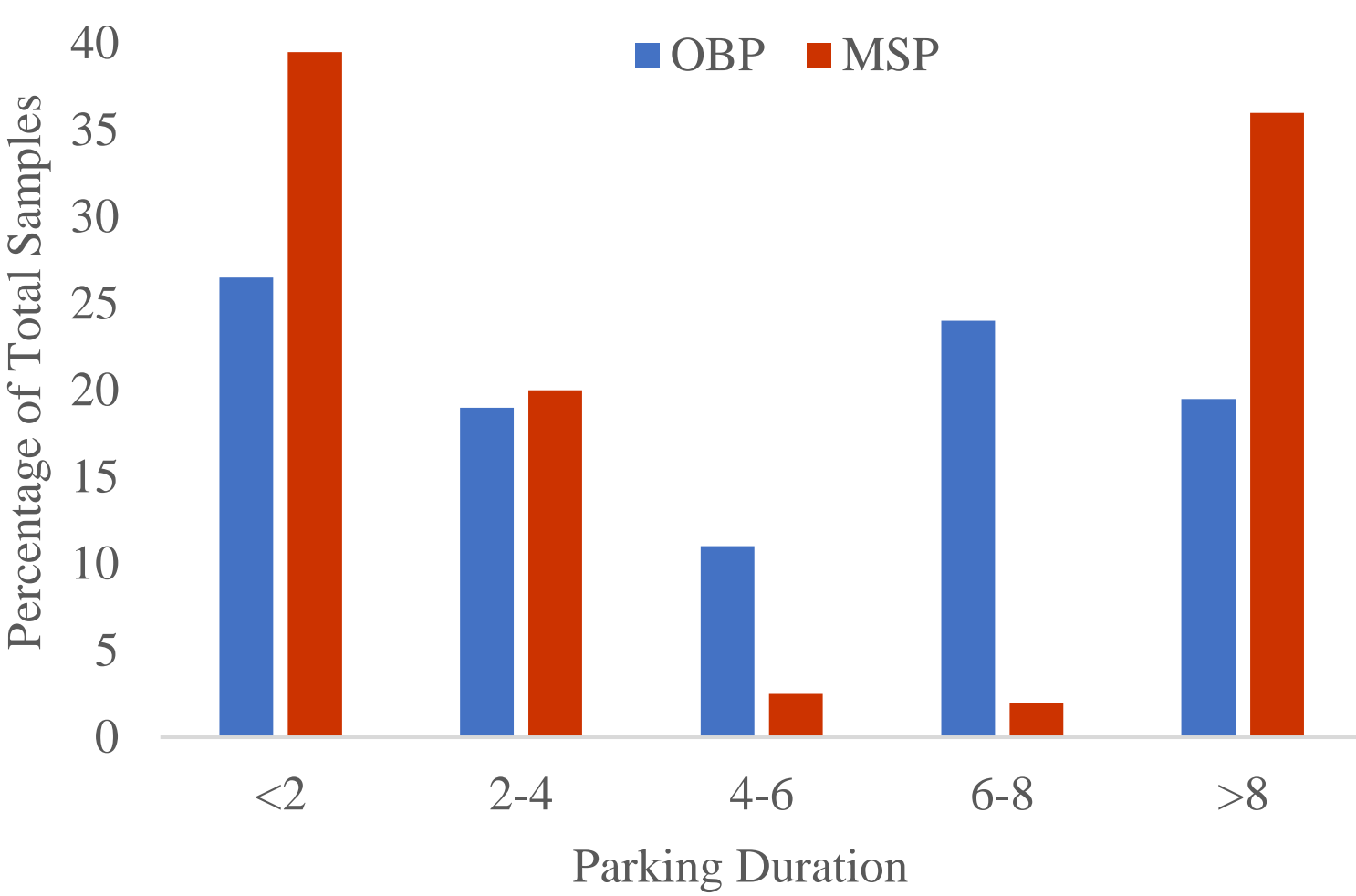

**Fig. 3.** Plot of Parking Duration Distribution

**Table 1.** Summary of Input Variables

| Variable | Levels | Descriptives (OBP) | Descriptives (MSP) |
|---|---|---|---|
| Personal Income (INR) | Below 20,000 | 0 % | 0.3 % |
| | 20,000 – 30,000 | 0.41 % | 13.6 % |
| | 30,000 – 45,000 | 24.4 % | 31.8 % |
| | 45,000 – 60,000 | 32.9 % | 35.4 % |
| | 60,000 – 80,000 | 25.1 % | 9.5 % |
| | Above 80,000 | 17.2 % | 9.5 % |
| Profession | Service | 60.8 % | - |
| | Business | 22.6 % | |
| | Student | 6..9 % | |
| | Self-employed | 9.6 % | |
| | Retired | 0.1 % | |
| | House-wife | - | |
| Number of Visits/month | Daily | - | 42.96 % |
| | 2-3 times a week | | 14.26 % |
| | Weekly | | 23.20 % |
| | Fortnight | | 8.76 % |
| | Occasionally | | 10.82 % |
| Purpose of Trip | Work | 63 % | 29.21 % |
| | Commercial | 28.5 % | 18.21 % |
| | Shopping | 8.3 % | 40.03 % |
| | Social | 0 % | 4.64 % |
| | Other | 0.2 % | 7.90 % |
| Travel Cost (INR) | - | 51.54±43.07 | 141.58±109.55 |
| Travel Time (min) | - | 37.52±27.34 | 67.77±42.93 |
| Parking fee/hr. (INR) | - | 19.23±9.10 | 17.83±6.25 |

Also, approximately 36% of the commuters came for work trip who park their cars for more than 6 hrs of duration. About 28% of total cars parked by the people whose trip purpose was business with average parking duration of 2-3 hrs. In case of MSP, 60% of the drivers parked their cars for lesser than 4 hrs. Approximately 50% of trips were found as shopping trips. Other major purposes of trip are work and commercial (nearly 40%) which are complimentary to the sellers and shopkeepers, and/or workers or managers working in big showrooms. Parking duration distribution shows that 59% of the cars parked for duration of 4 hrs or less and nearly 31% for a duration of 8 hrs or more. These figures clearly distinguish between parking by the shoppers and sellers. Descriptives show that travel cost and travel time are higher in case of MSP compared to OBP i.e., the service area for those markets is larger. During the interviews, it was found that most people are using private vehicles because the fare for travel by metro is higher than that is by car. Taking an example from the interview, four persons come for shopping would pay a total fare of 4×35=140 INR, while the travel cost + parking cost might be lower, say 50+60=110 INR (60 INR for 3 hours of parking). Hence, it would be cheap to use private car. Besides, it can be seen that the parkers are paying nearly equal parking fees for both the cases.

## 5. Analysis

### 5.1. ANN model

The multilayer perceptron BPNN was considered to model parking duration. Before the model training, the dataset was pre-processed and standardized using *center-scale* method which rescales the data to have a mean of 0 and a standard deviation of 1 (unit variance). The sigmoid activation function was considered on both sides of the hidden layer during the network learning process. To get the optimize neural network architecture and to stop the training, cross-validation was used with *hidden layer-size* and *weight-decay* parameters. Initial size of hidden layer was considered as 1:20 (i.e., train() considers each case with hidden layer size vary from 1 to 20). Decay parameter was chosen from 0, 0.001, 0.01 and 0.1, depending on which model depicts the best accuracy. The model with highest accuracy (i.e., minimum error) and Cohen's kappa value (Sim and Wright, 2005) was used for further analysis and prediction. The kappa was calculated based on the confusion matrix created for training dataset and predicted parking duration classifications (See Appendix). In order to test the applicability of LIME on the developed BPNN model, 6 samples at random were excluded from the training dataset. The LIME interpreter was asked to predict correct classification for the given data of these 6 samples and the same was compared with the actual response. Remaining dataset was used to train the neural network model. The subsequent sections discuss the model performance as well as applications of Garson's algorithm and LIME algorithm for interpretation of the developed model.

### 5.2. Office-Business land-use

As stated earlier, six parameters viz. personal income, profession, trip purpose, parking fee/hr, travel cost and travel time were included in model development. The structure of the developed model was: input layer, consist six parameters as depicted above (20 neurons), one hidden layer with 4 neurons and one output layer with five classifications of parking duration. The train() function took approximately 20903 iterations to select best optimized model for a given data. The combinations of *size* and *decay* parameters with respect to accuracy, and the final optimized model is shown Fig. 4. It illustrates the highest model accuracy of 91.4% with 4 hidden neurons and 0.01 weight decay. The confusion matrix (See Appendix) depicts the *observed* prediction accuracy of 91.91% against an *expected* accuracy of 21.38% come up with the kappa of 0.89. It implies substantially good neural network model, which can be further used to predict the target variable.

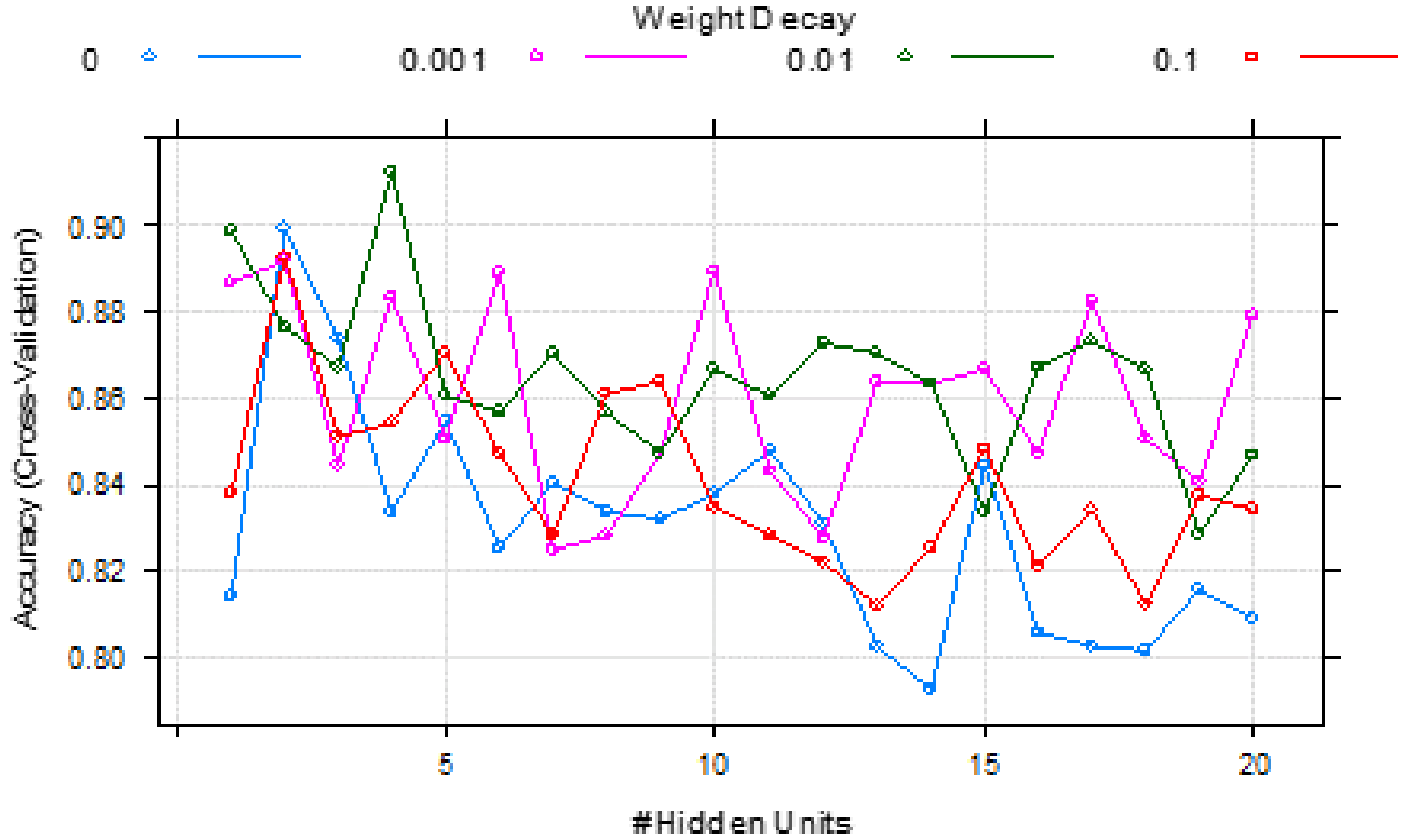

**Fig. 4.** Plot of model accuracy with tuning parameters

### 5.2.1. Model Interpretation and Prediction

As stated earlier, the Garson's algorithm and LIME algorithm were used for the interpretation of BPNN model. In R, garson() function from the library package *NeuralNetTools* was employed to determine the relative importance of each predictor in the model. Table 2 demonstrates the relative importance of input variables in the parking duration model. The results suggest that parking fee/hr is the most important predictor for a parking duration outcome, followed by profession and income.

Though the results from Garson's algorithm provide insights of predictor's importance in the model, it cannot be identified which parameter has positive influence and which contradicts the outcome variable. Moreover, the relative importance value cannot be implemented to further predict/classify the dependent variable for a given unseen dataset. Therefore, the use of the algorithm is limited only to analyse the weights of predictors in model. This limitation can be overcome by application of LIME.

**Table 2.** Relative importance using Garson's algorithm

| Variable | Relative Importance |
|---|---|
| Income_30k-45k | 0.075 |
| Income_45k-60k | 0.088 |
| Income_60k-80k | 0.058 |
| Purpose_Shopping | 0.075 |
| Purpose_Work | 0.038 |
| Profession_Self-emp | 0.027 |
| Profession_Service | 0.117 |
| Profession_Student | 0.056 |
| Travel time | 0.034 |
| Travel cost | 0.070 |
| Parking fee/hr | 0.361 |

The interpretation of complex models is possible using LIME which provides a qualitative link between the predictors and response. The *agnostic* descriptor states that it can be used to provide insight into a black box process and approximates the response locally. In this study, six data-points unseen during training phase were used to check the interpretability of model and competency of LIME in predicting them correctly.

The lime() function is the principal function which looks for the training dataset used for to develop model and the neural network model that need to be explained. The lime() returns an object called explainer that is passed to the function explain(). It takes new unknown dataset along with the explainer and returns a matrix with prediction explanations. This function also reports the class probability for the predicted response. Higher the probability, more the confidence that prediction be identical to observed classification. To assure this, six observations from the main dataset were excluded to examine the predicted classification with that of the observed. Fig. 5, called *feature plot,* demonstrates the prediction results for the dataset obtained from OBP.

In the plot, *case* shows the sample number, *label* indicates the predicted value of outcome, *probability* is the predicted probability of the label and *explanation fit* measures the quality of model used for the explanation. The feature plot demonstrates the predication probability which varies from 0.88 to 1. The performance, in terms of classification prediction, obtained by LIME clearly depicts the efficacy of LIME in predicting outcome i.e., parking duration. The prediction probability of each outcome observed to be high which augment the applicability of developed ANN model. Each of the six outcomes greatly influenced by the parking fees. The extensive use of the plot is discussed in detail in the subsequent section.

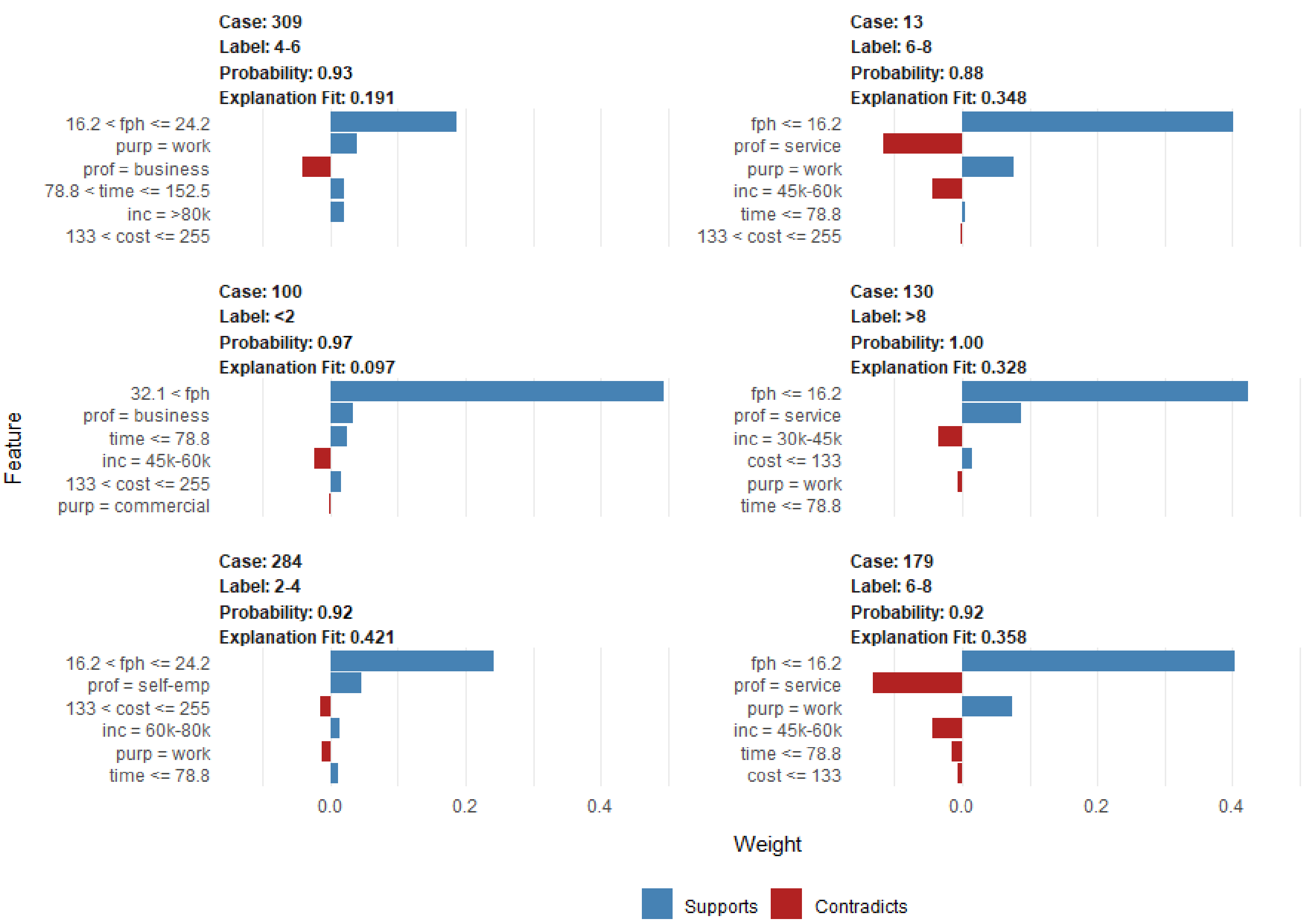

**Fig. 5.** Feature plot produced by plot_features() function showing the parameter influence
(**Note:** fph is parking fee/hr.)

## 5.3. Market/Shopping land-use

In the parking duration model for shopping area, number of visits per month is included in place of profession in the neural network model for OBP land-use. An optimized model was developed after 22131 iterations after tuning the *size* and *decay* parameters, which is shown in the Fig. 6. The final model structure includes 20 input neurons, 8 hidden neurons and 5 output neurons. The final model accuracy during the training was 92.89% with 8 hidden units and 0.1 weight decay. The Cohen's kappa value was estimated as 0.90 based on the *observed* accuracy of 93.18% and the *expected* prediction accuracy of 32.29% (See Appendix). It justifies the efficiency of the developed neural network model in predicting the response variable (i.e., parking duration) with significantly higher accuracy. The next subsection describes the model interpretation and the predictability of the same.

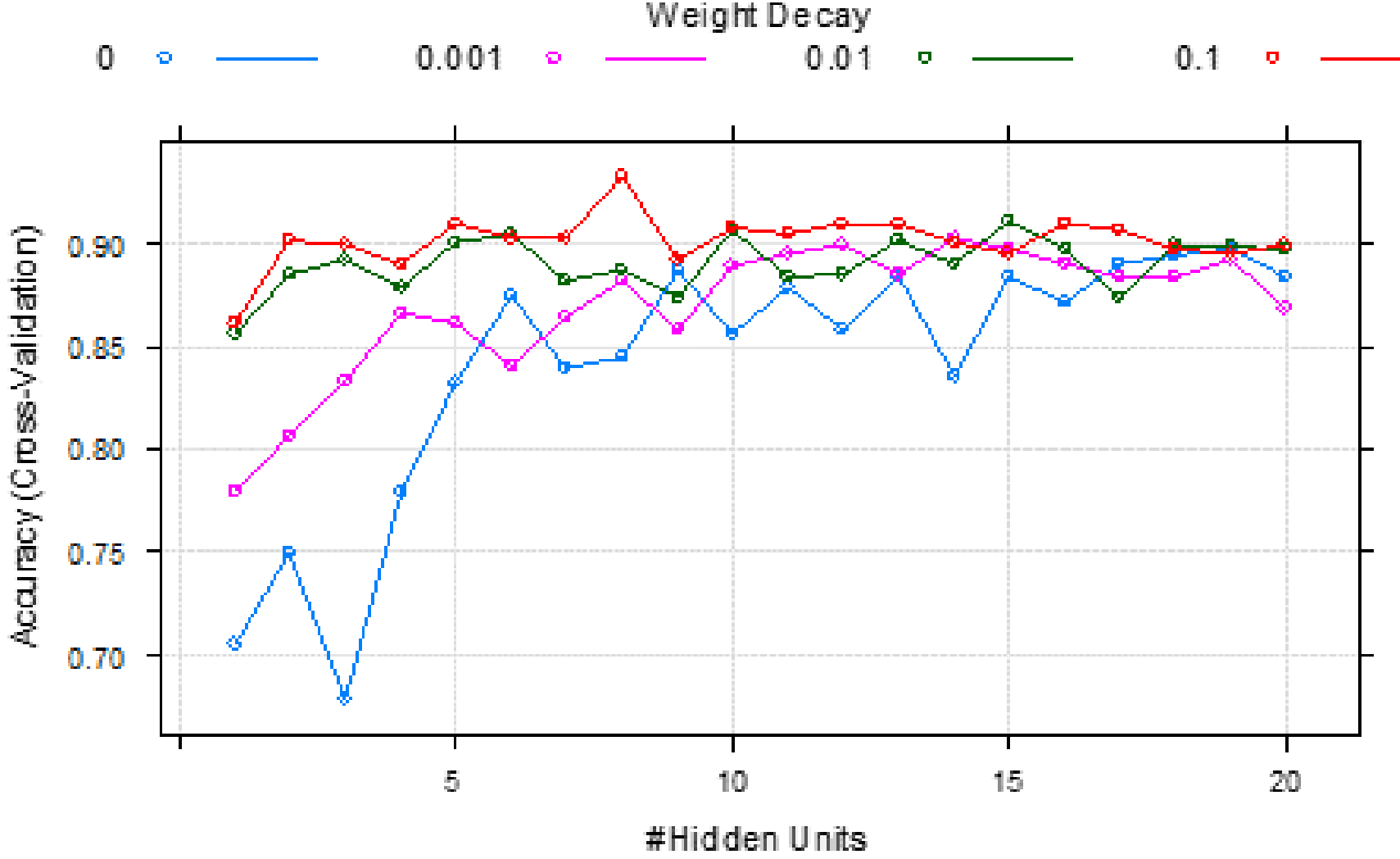

**Fig. 6.** Plot of model accuracy with tuning parameters

### 5.3.1. Model Interpretation and Prediction

For the MSP dataset, the same procedure was followed as illustrated for OBP land-use data. The Table 3 depicts the relative importance of the parameters that was to be significant in the neural network model. It can be seen that the parking fee/hr. has the highest influence on parking duration. In addition, Purp_Shopping (shopping as a purpose of trip) greatly influenced the parking duration as this model was developed particularly for the MSP. To interpret the ANN black-box and predict the parking duration based on the considered parameters, 6 unseen samples were incorporated to predict using LIME algorithm. Fig. 7 shows the feature plots for the randomly selected 6 samples. The results portray the excellent prediction ability of LIME as the probabilities of predicting all six cases are higher than 90% with same label which was observed from the survey data. It can be observed that parking fees/hr has negative effect on a higher parking duration (i.e., 6-8 and >8 hrs.) while it presents positive trend for parking duration lesser than 6 hrs.

**Table 3.** Relative importance using Garson's algorithm

| Variable | Relative Importance |
|---|---|
| Visit_Fortnight | 0.046 |
| Visit_Occassionally | 0.055 |
| Visit_Two-three times | 0.047 |
| Visit_Weekly | 0.043 |
| Inc_20k-30k | 0.043 |
| Inc_30k-45k | 0.062 |
| Inc_45k-60k | 0.071 |
| Inc_60k-80k | 0.077 |
| Inc_>80k | 0.051 |
| Purp_Other | 0.070 |
| Purp_Shopping | 0.091 |
| Purp_Social | 0.054 |
| Purp_Work | 0.043 |
| Travel time | 0.076 |
| Travel cost | 0.076 |
| Parking fee/hr | 0.096 |

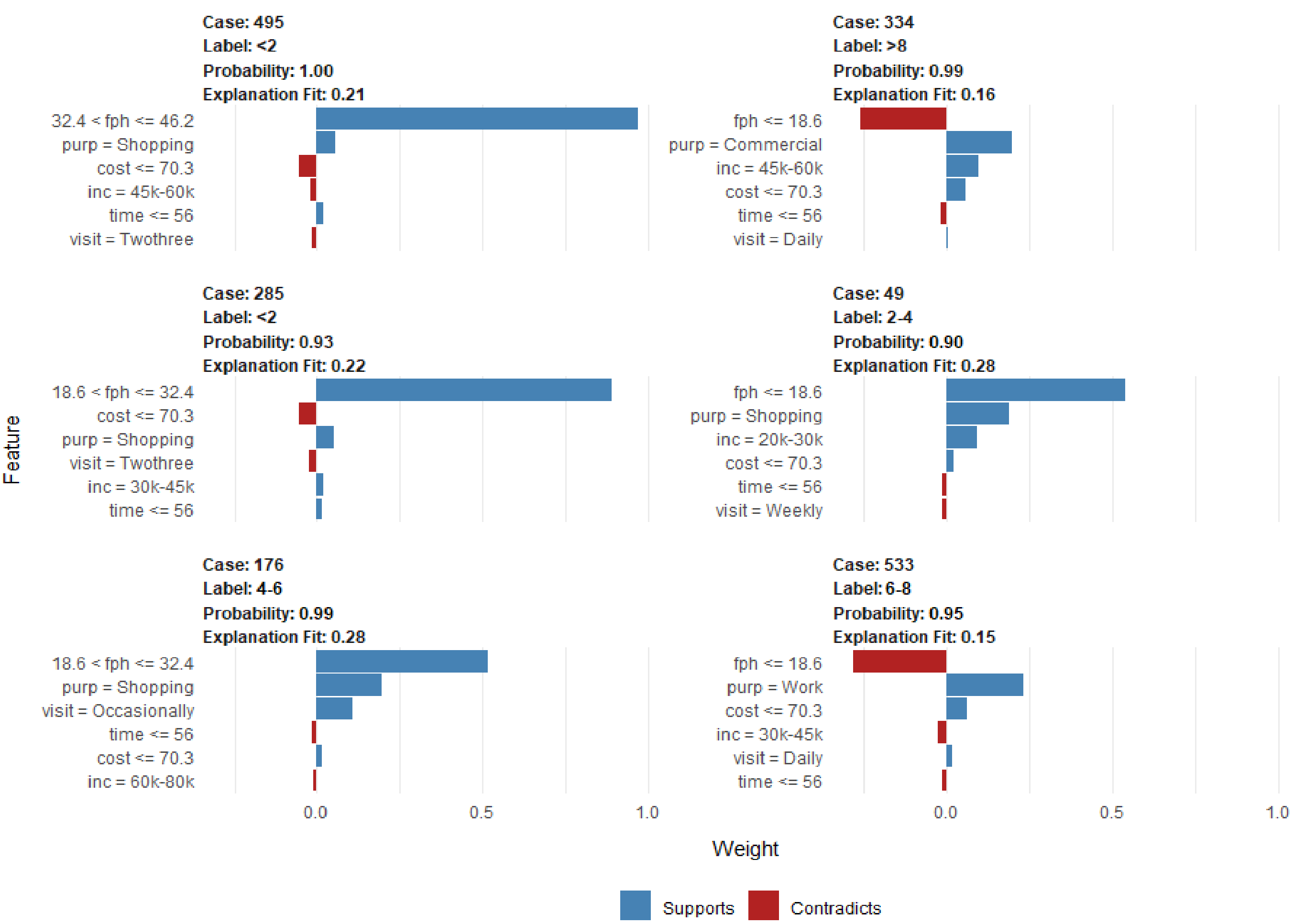

**Fig. 7.** Feature plot produced by plot_features() function showing the parameter influence (**Note:** fph is parking fee/hr.)

## 6. Discussions

This section discusses the behavioural response based on the results and the possible policy implications in developing countries like India. Results yielded by Grasons's algorithm and LIME parking duration in OBP are shown in Section 5.2. It can be seen that the results produced by LIME are competent since prediction probabilities are very high and approving the real-data. The results developed by the LIME are in line with the previous studies examining the parking demand in terms of duration-choice (Zong and Wang, 2015; Zong et al., 2019). But the previous studies did not include how well the model will perform on an unseen data which limits its applicability while this study demonstrates the higher prediction probabilities on the new data. It reflects the suitability of the model to be pertinent to other regions as well. Moreover, the model provide insights on each observation invidually how the selected variables influence driver's decision which outperforms the previous models.

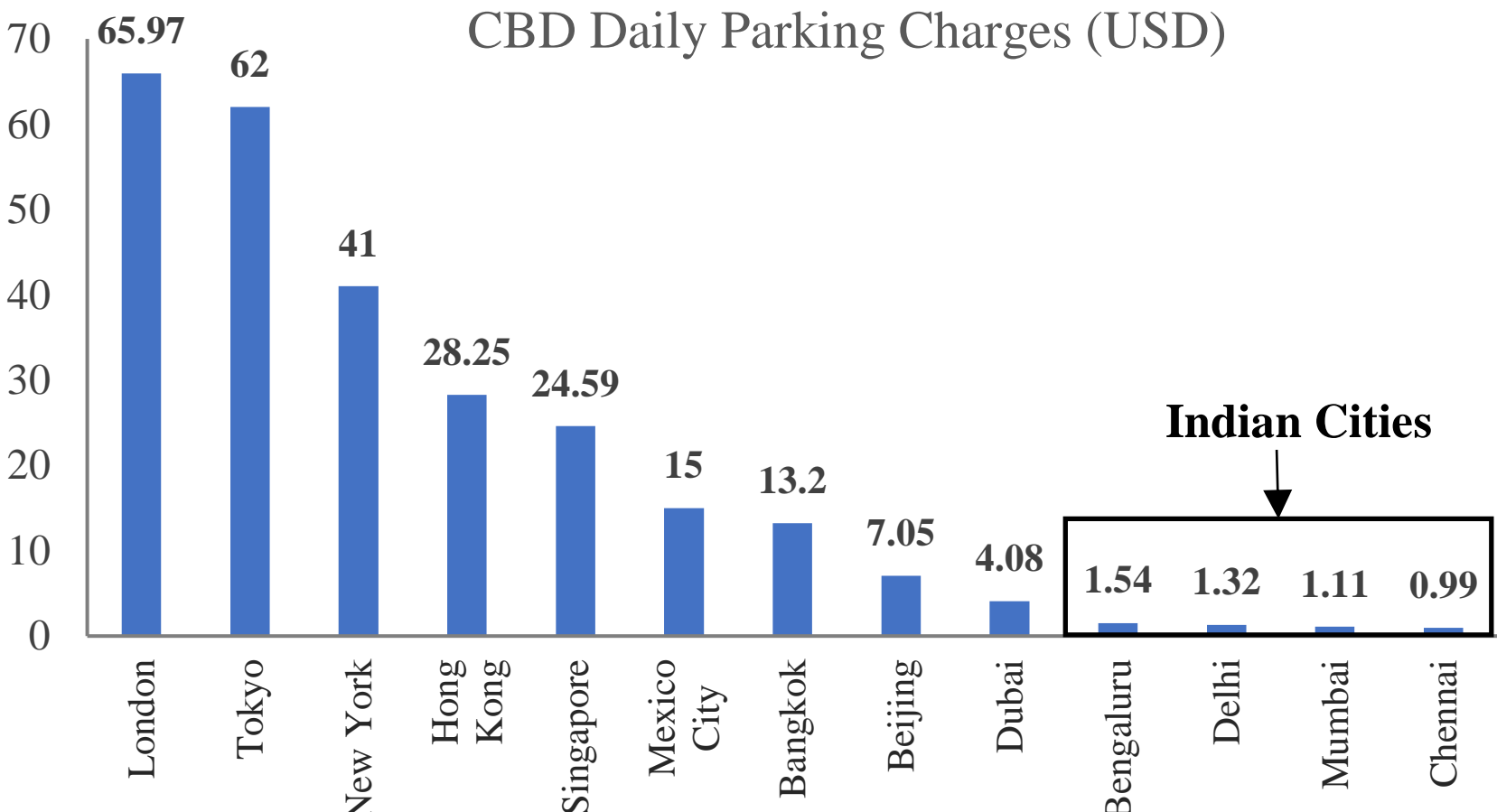

**Fig. 8.** Daily Parking Charges in cities across the Globe

As the outcomes of this study suggest that the parking fee is the most important parameter while deciding the parking duration, it should be consider as a target parameter in policy implications. Further, the profession and income significantly contribute to parking duration model. The results provide the information regarding the target group of parkers and the target variable which is useful while forming the parking policies for effective and sustainable parking management. For OBP, Prof_Service, Purp_Work and Income group of 30k-60k INR seem to be the traget groups as they are greatly influencing the long-term parking duration (Fig. 4). It means the drivers with income between 30k-60k INR commuting by car are vulnerable to change in parking charges. As the authorities hold the statistics of patrons, they can frame efficient policies with this group keeping in mind. According to the Handbook of Urban Statistics (2019), Indian parking fees are the lowest in the world (see Fig. 8). The peculiarity is that the parkers pay next to nothing to use valuable lands even in CBDs, nor are these costs recovered through taxes, which encourages more and more people to use private vehicles. This amount will work out to be much larger if the land-cost and investments for parking in prime areas are taken into account. As this study provides an insight which group of people will use the parking space for how much duration, the policymakers can target those by managing those lots with suitable parking charges. As mentioned earlier, about 25% of the OBP is employer-paid but that cost to them much lesser than commuter's willingness-to-pay which implies welfare loss. Furthermore, the free/subsidized parking encourages commuting by car. As per the cases from United States and Canada, employer-paid parking boosts the number of commuting cars by about 36% and the share of car-based commuting by 60% (Shoup, 2005a). Hence, an appropiate policy and parking fare structure should be framed to manage the parking supply as well as minimum parking requirements. It may include Parking cash-outs (Shoup, 2005b), Space restraint policy (MoHUA, 2014), Performance based and Dynamic parking pricing (Parmar et al., 2020). In order to maintain social optimum, commuters using private vehicle should pay their own parking fees. In switching from employee-paid parking to employer-paid parking, commuters vividly shift from public to private mode. Subsequently, it requires additional costly land in CBDs for inefficient and under-productive parking lots and ultimately reduce the urban utility level (Brueckner and Franco, 2018). In this case, appropiate increase in parking fare or parking cash-out policy should be mandated particularly, in coutries like India to effectively restore the efficiency of parking lots and the urban eqillibrium.

For the MSP, it is noted that the income, trip-purpose and parking fees are the most influential in predicting parking duration. Moreover, travel cost is affected upto a certain level for higher parking duration which probably describes the cost comparision shown in the Subsection 4.1. It is observed that the parking fees are more important to short-term parkers in comparision to the long-term parkers. Besides, it contradicts the parking demand (i.e. duration) for long-term parking which means that they are less liable to affect from the changes in parking price policies. These results come when all the study locations are well connected by means of public bus and mtero which should be taken seriously. Apperently, the long-term parkers dominate the parking accumulation. Consequently, the graded parking price structure based on the parking duration (i.e., higher fees for longer duration) can be formulated stringently which may shift long-term parkers to public mode or short-term parking (other than shop-keepers). It would increase the parking efficiency level in addition to relief to saturated capacity. Taking into cosideration prescribed cost comparision, the hourly-demand based parking fare structure can be imposed as a suitable policy alternative. Besides, the proper time-varying fee structure will ensure the high parking occupancy rates and prevent the overloaded demand in an efficient manner (Small and Verhoef, 2007). In line with the current study, a differentiation can be predicted between long-term and short-term parking considering various factors accroding to which the two-fold policy can be framed. Study in Beijing shows that the parking fees increase from 0-0.15 USD/hr to 0.8-1.6 USD/hr lead to increase in the probability of parking for less than three hours from 40.85% to 92.61% (Zong and Wang, 2015). In addition, probabilities of parking for 3-12 hours and >12 hours decrease from 33.3% to 5.04% and from 25.85% to 2.35% respectively.

During the interview survey, it was revealed that the parkers seem to adjust their parking duration (probably the activity duration) conferring to where they park instead of planning ahead of arriving at the destination. From human psychology perspective, this suggests if they find cheap parking space, they park for a loger duration! In general, choosing a parking duration follows the location of the parking also (i.e., land-use and parking fees) and the both are affected by the time of parking. It suggests that policy interventions targeting in one parking-oriented decision may indirectly influences the other parking-related decisions. If transport policy aimed at encouraging the off-peak period parking, it could increase the probability of on-street parking (which may be negative as far as CBD concerns) and reduce a long-term parking (which is positive from efficiency and sustainibility point of view). Hence, the impacts of all three scenarios should be taken into consideration concurrently while framing parking policies.

## 7. Conclusions

Urban areas in the developing coutries are dominated by private automobile riders due to inefficient tranport policies and declining public transport. As a result, parking became one of the major issues across the sustainable development. An effective parking management may not only the travel behaviour and mode choice but also potentially improve the economic vitality of an area by imrpoving the turnover. In this context, as attempt is made to develop an efficient model which can predict one of the parking demand measures i.e., parking duration for two types of land-uses. To evaluate the contribution of various parameters in parking duration prediction, the ANN model has been developed as decribed in previous section. Besides, the importance of the parameters has been calculated using the Garson's partitioning of weights algorithm. As this algorithm has limitations in estimating the influence of contributing variables on response variable, a recently developed LIME algorithm was used to predict the parking duration and to evaluate *nature of contribution* to the prediction. It was observed that the economic parameters are mostly dominated in the model besides the nature of trip. Finally, policy-based discussion is framed with consideration of the target groups and target variable. A parking price-based regulatory framework is dicussed which is influential for the well-organized parking management system. This study is extremey helpful and ubiquitous in nature in framing strategies and policies ruminating diversity in the nature of target groups in rapidly growing urban world. As the literature lack in sufficient information on how people select a parking duration for a given trip, it should be examined how people change (or not) their parking duration considering comprehenssive set of parameters including accessibility to parking, spatial characteristics in the vicinity of parking lots, different set of parking price structures, etc. in future.

## Appendix: Confusion matrices from optimized models

**Table 4.** Confusion Matrix for OBP model

| Predicted Duration→<br>Observed Duration↓ | < 2 | 2-4 | 4-6 | 6-8 | > 8 | Total Observed |
|---|---|---|---|---|---|---|
| < 2 | 116 | 6 | 1 | 0 | 0 | 123 |
| 2-4 | 8 | 60 | 2 | 0 | 0 | 70 |
| 4-6 | 5 | 5 | 53 | 3 | 0 | 66 |
| 6-8 | 0 | 0 | 2 | 110 | 3 | 115 |
| > 8 | 0 | 0 | 0 | 4 | 104 | 108 |
| Total Predicted | 129 | 71 | 58 | 117 | 107 | 482 |
| Accuracy | Correctly classified = 443 misclassified = 39 | | | | | |
| | Percent Accuracy = 91.91% | | | | | |

**Table 5.** Confusion Matrix for MSP model

| Predicted Duration→<br>Observed Duration↓ | < 2 | 2-4 | 4-6 | 6-8 | > 8 | Total Observed |
|---|---|---|---|---|---|---|
| < 2 | 255 | 25 | 2 | 0 | 0 | 282 |
| 2-4 | 9 | 108 | 3 | 1 | 0 | 121 |
| 4-6 | 0 | 5 | 19 | 1 | 0 | 25 |
| 6-8 | 0 | 0 | 0 | 12 | 0 | 12 |
| > 8 | 0 | 0 | 0 | 0 | 235 | 235 |
| Total Predicted | 264 | 138 | 24 | 14 | 235 | 675 |
| Accuracy | Correctly classified = 629 misclassified = 46 | | | | | |
| | Percent Accuracy = 93.18% | | | | | |